\documentclass[letterpaper,twocolumn,fleqn]{article} 

\usepackage{ist}
\usepackage{algorithm}
\usepackage{array}
\usepackage{amsmath}
\usepackage{amssymb}
\usepackage[utf8]{inputenc}
\usepackage[export]{adjustbox}
\usepackage{textcomp}
\usepackage{stfloats}
\usepackage{csquotes}
\usepackage[super]{nth}
\usepackage[skip=0.5\baselineskip]{caption}
\usepackage{url}
\usepackage{verbatim}
\usepackage{siunitx}
\usepackage[font={footnotesize}, textfont=sf]{subfig} % for subfigures
\usepackage{relsize}
\usepackage{tikz}
\usepackage{graphicx}
\usepackage[citebordercolor=white, linkbordercolor=white, urlbordercolor=black, hidelinks, pdfborderstyle={/S/U/W 1}, breaklinks]{hyperref}
\usetikzlibrary{shapes.geometric, arrows, fit,matrix, backgrounds,positioning,calc, decorations.pathreplacing,calligraphy}
\title{Measuring Natural Scenes SFR of Automotive Fisheye Cameras}

\author{Daniel Jakab$^1$, Eoin Martino Grua$^1$, Brian Micheal Deegan$^2$, Anthony Scanlan$^1$, Pepijn Van De Ven$^1$,
and Ciarán~Eising$^1$\\
$^1$Dept. of Electronic and Computer Engineering, University of Limerick, Castletroy, Co. Limerick  V94 T9PX, Ireland\\
$^2$Dept. of Electrical and Electronic Engineering, University of Galway, Galway H91 TK33, Ireland}
 
\vspace{-1.2cm}
\date{} % date has an empty field.

\begin{document} 

\maketitle

\thispagestyle{empty} % prevents the first page to be numbered

%%%%%%%%%%%%%%%%%%%%%%%%%%%%%%%%%%
% Abstract
%%%%%%%%%%%%%%%%%%%%%%%%%%%%%%%%%%

\begin{abstract}
The Modulation Transfer Function (MTF) is an important image quality metric typically used in the automotive domain. However, despite the fact that optical quality has an impact on the performance of computer vision in vehicle automation, for many public datasets, this metric is unknown. Additionally, wide field-of-
view (FOV) cameras have become increasingly popular, particularly for low-speed vehicle automation applications. To investigate image quality in datasets, this paper proposes an adaptation of the Natural Scenes Spatial Frequency Response (NS-SFR) algorithm to suit cameras with a wide field-of-view.
\end{abstract}

\section{Introduction}
\label{sec:intro}
Wide field-of-view (FOV) cameras are a unique optical and sensor combination for low-speed vehicle automation \cite{Kumar2023, Eising2021}. The interest in this sensor modality has spawned several datasets \cite{Yogamani2019, eichenseer2016data, sekkat2022synwoodscape}, which encourages the development of perception algorithms. However, the optical quality of such datasets is unknown, despite the fact that the connection between optical quality and vision performance has been demonstrated \cite{muller2022simulating}.

%Optical Effects affect optical quality in camera lenses and are crucial to automotive.
In \cite{muller2022simulating} an obvious impact of optical lens quality is seen on computer vision performance, where a Cooke Triplet lens degraded the performance of object detection on a BDD100k dataset using a sample of 10,000 images. The BS ISO12233:2017 standard \cite{ISOBS2017} and Slanted Edge algorithm \cite{burns2018camera} were used to measure optical quality across the spatially variant domain of the camera. In this experiment, test charts were used to measure the optical quality of a camera using three different lens blur models (Gaussian, Superposition, and isoplanar patches). This study was limited to only one lens. This suggests that more information on optical quality is required for publicly available datasets. As of this moment, no datasets provide a means of measuring optical performance using test charts, which suggests replicating optical quality measurements from \cite{muller2022simulating} is difficult and time-consuming.
As a strategy to measure image quality performance in the publicly available datasets, the Natural Scenes Spatial Frequency Response (NS-SFR) developed by van Zwanenberg et al. specializes in extracting slanted edges from natural scenes \cite{van2023tool, van2022camera, van2021natural, van2018estimation}. We adapt NS-SFR to various datasets of spatially variant FOV.

In this paper, we introduce a modified pipeline for generalizing the NS-SFR algorithm to both narrow and wide FOV lenses. The original NS-SFR is unable to eliminate areas not part of the natural scene, which biases optical quality measurements. This especially applies to automotive wide FOV cameras which have ego-vehicle occlusion and lens aperture in the images. Thus, we propose the following contributions in this paper:
\begin{enumerate}
    \vspace{-0.2cm}
    \item{We propose \textbf{regional masking around the scenes} in wide FOV images to remove ego-vehicle occlusion and mechanical vignetting},
    \item{We propose \textbf{a valid and invalid Region of Interest (ROI) Selection System} which chooses edges with the least amount of edge enhancement and noise of slanted edges},
    \item{We propose using \textbf{an Adaptive Radial Distance Analysis} for evaluating optical measurements where radial segments adapted to the regional mask of scene},
    \item{First analysis across \textbf{multiple dynamic automotive scenes} where each scene has a unique camera lens},
    \vspace{-0.2cm}
\end{enumerate}
Our experiments are focused on four datasets: three are real-life datasets with 90°, 185°, and 190° FOV respectively, and the \nth{4} dataset is based on simulation with 190° FOV.

%In the following sections of this paper, we will discuss: (1) the context of NS-SFR, (2) the methodology, (3) show experimental results, and (4) a discussion and further experiments to consider.
%\section{Related Work}
%Briefly discuss previous experiments performed by Oliver on the DSLR \& smartphone cameras.
\section{Context of NS-SFR}
For this paper, we use the BS ISO12233:2017 \cite{ISOBS2017}  Slanted Edge algorithm as a basis for experiments where the adapted NS-SFR focuses on finding edges on both narrow and wide FOV natural scenes. As far as we know, these experiments are the first attempt at Edge Spatial Frequency Response (e-SFR) for wide FOV cameras without using test charts.  %Test charts with predefined measurement locations are used for distorted camera lens measurements. Test charts available on the Imatest website are only applicable to $\leq \pi$ or $\leq \ang{180}$ FOV cameras.
More sophisticated cameras exceeding \ang{180} FOV require test charts with greater distortion. There has been little research on taking natural scenes as part of the measurement process for automotive scenes. %NS-SFR proves useful in this regard as it measures slanted edges using MTF in Regions of Interest (ROIs) on a canny edge detector of a scene \cite{van2019edge}.  
%MTF is a common factor measuring image sharpness, one of the most important parameters in image quality. %measures the sharpness of an image creating a distinct graph where blurring in an image is the attenuation of high-frequency components. In cameras with large apertures and short focal lengths, details in the background are difficult to maintain as is the case with wide FOV or fisheye cameras.

\section{Methodology}
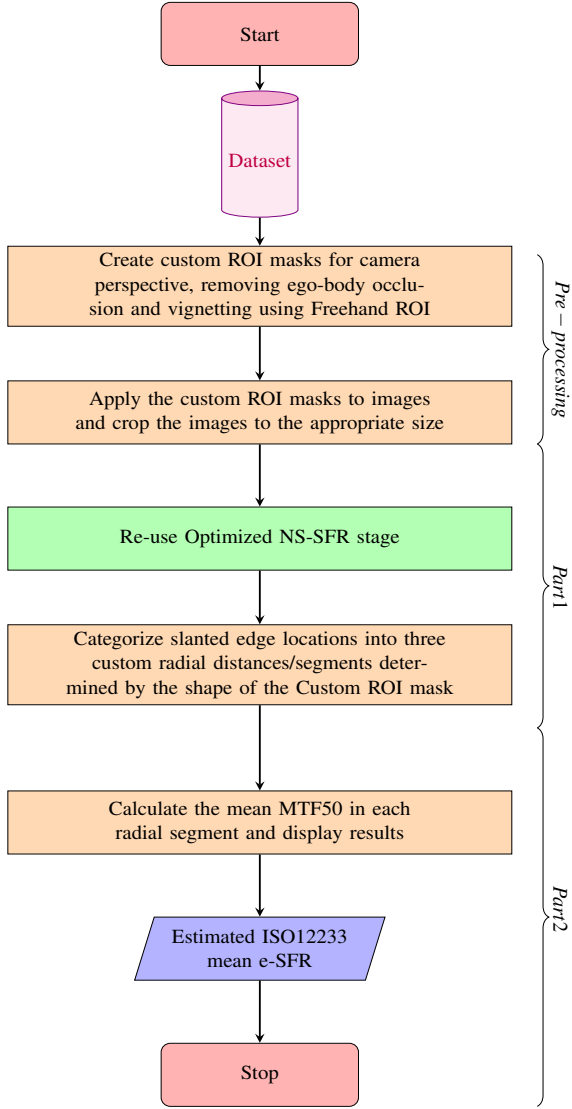
\begin{figure}[t]
    \centering
    \captionsetup{font={normalsize}, textfont=sf}
    \tikzset{
    use page relative coordinates/.style={
        shift={(current page.south west)},
        x={(current page.south east)},
        y={(current page.north west)}
    },
}
\tikzstyle{startstop} = [rectangle, rounded corners, 
minimum width=3cm, 
minimum height=1cm,
text centered, 
draw=black, 
fill=red!30]

\tikzstyle{comp} = [trapezium, 
trapezium stretches=true, % A later addition
trapezium left angle=70, 
trapezium right angle=110,
text width=3cm,
minimum width=3cm, 
minimum height=1cm, text centered, 
draw=black, fill=blue!30]

\tikzstyle{pers} = [trapezium, 
trapezium left angle=70, 
trapezium right angle=110,
inner sep=12pt,
text width=3mm,
minimum width=2cm, 
minimum height=1cm,
draw=black, fill=blue!30]

\tikzstyle{database} = [cylinder, 
    draw = violet, 
    text = purple,
    minimum height=2cm,
    cylinder uses custom fill, 
    cylinder body fill = magenta!10, 
    cylinder end fill = magenta!40,
    aspect = 0.2, 
    shape border rotate = 90]

\tikzstyle{process} = [rectangle, 
minimum width=7.45cm, 
minimum height=1cm, 
text centered, 
text width=7.45cm, 
draw=black, 
fill=orange!30]

\tikzstyle{reuse} = [rectangle, 
minimum width=7.45cm, 
minimum height=1cm, 
text centered, 
text width=7.45cm, 
draw=black, 
fill=green!30]

\tikzstyle{arrow} = [thick,->,>=stealth]

\resizebox{7.7cm}{14.7cm}{
    \begin{tikzpicture}[node distance=2cm]
    \node (start) [startstop] {Start};
        \node (db1) [database, below of=start]  at (0,0){Dataset};
        \node (prep1) [process, below of=db1] {Create custom ROI masks for camera  perspective, removing ego-body occlusion and vignetting using Freehand ROI};
        \node (proc1) [process, below of=prep1] {Apply the custom ROI masks to images and crop the images to the appropriate size};
        \node (proc2) [reuse, below of=proc1] {Re-use Optimized NS-SFR stage};
        \node (proc3) [process, below of=proc2] {Categorize slanted edge locations into three custom radial distances/segments determined by the shape of the Custom ROI mask};
        \node (proc4) [process, below of=proc3, yshift=-0.5cm] {Calculate the mean MTF50 in each radial segment and display results};
        \node (out1) [comp, below of=proc4] {Estimated
ISO12233 mean e-SFR};
        \node (stop) [startstop, below of=out1] {Stop};

        % Simple brace
        \draw [decorate,
            decoration = {brace,mirror, raise=120pt, amplitude=5pt}] (0,-6.5) --  (0,-3.5)
            node[pos=0.9, right=130pt,black, rotate=-90]{$Pre-processing$};
        \draw [decorate,
            decoration = {brace,mirror, raise=120pt, amplitude=5pt}] (0,-11) --  (0,-6.5)
            node[pos=0.6, right=130pt,black, rotate=-90]{$Part 1$};
        \draw [decorate,
            decoration = {brace,mirror, raise=120pt, amplitude=5pt}] (0,-17) --  (0,-11)
            node[pos=0.6, right=130pt,black, rotate=-90]{$Part 2$};
        
        \draw [arrow] (start) -- (db1);
        \draw [arrow] (db1) -- (prep1);
        \draw [arrow] (prep1) -- (proc1);
        \draw [arrow] (proc1) -- (proc2);
        \draw [arrow] (proc2) -- (proc3);
        \draw [arrow] (proc3) -- (proc4);
        \draw [arrow] (proc4) -- (out1);
        \draw [arrow] (out1) -- (stop);
    \end{tikzpicture}
}
    
    \vfill
    \caption{Adapted NS-SFR.}
    \label{fig:nssfr-adapt}
    \vspace{-0.3cm}
\end{figure}

The proposed Automotive NS-SFR pipeline is depicted in Fig.\ref{fig:nssfr-adapt}. As the goal is to measure the optical performance of a single camera, the dataset must contain images from only one camera. For wide FOV cameras, results are biased where NS-SFR selects the edges that are not part of the natural scene. To avoid result bias, regional masking is used and is applied to the dataset only if parts of the vehicle or camera aperture cover parts of the natural scene, biasing the results. Next, the optimized NS-SFR \cite{van2023tool} is applied to the dataset for ROI selection as illustrated by the green box in Fig. \ref{fig:nssfr-adapt}. The ROI locations of the slanted edge measurements were categorized into three spatial areas on the images: (1) center, (2) middle, and (3) edge. Finally, the mean e-SFR of each radial segment was calculated.
\subsection{Regional Masks}
\begin{figure*}[t]
    \centering
    \captionsetup{font={normalsize}, textfont=sf} 
    \hspace{-0.55cm}
    \subfloat[KITTI\label{fig:KITTI}]{\includegraphics[width=2.5in, keepaspectratio]{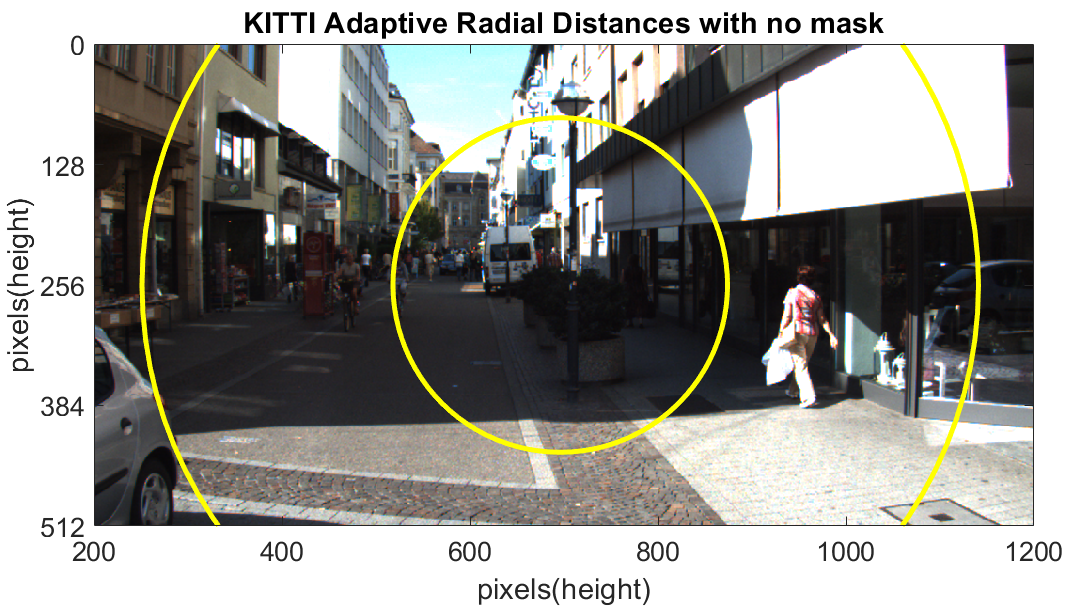}}
    \hspace{1.2cm}
    \subfloat[LMS\label{fig:lms}]{\includegraphics[width=2.2in, keepaspectratio]{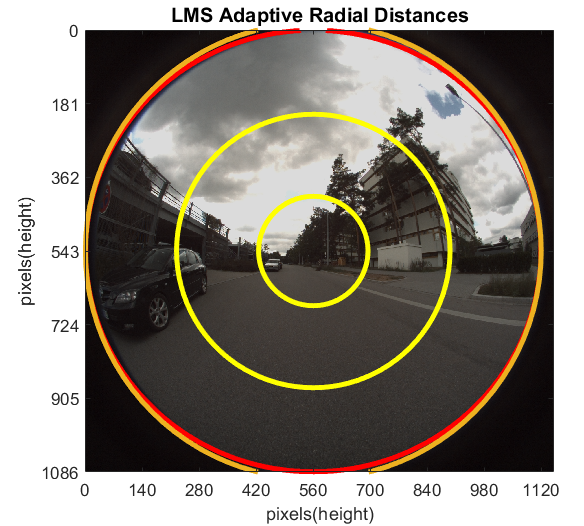}}\\
    \hspace{0.1cm}
    \subfloat[Woodscape\label{fig:wood}]{\includegraphics[width=2.8in, keepaspectratio]{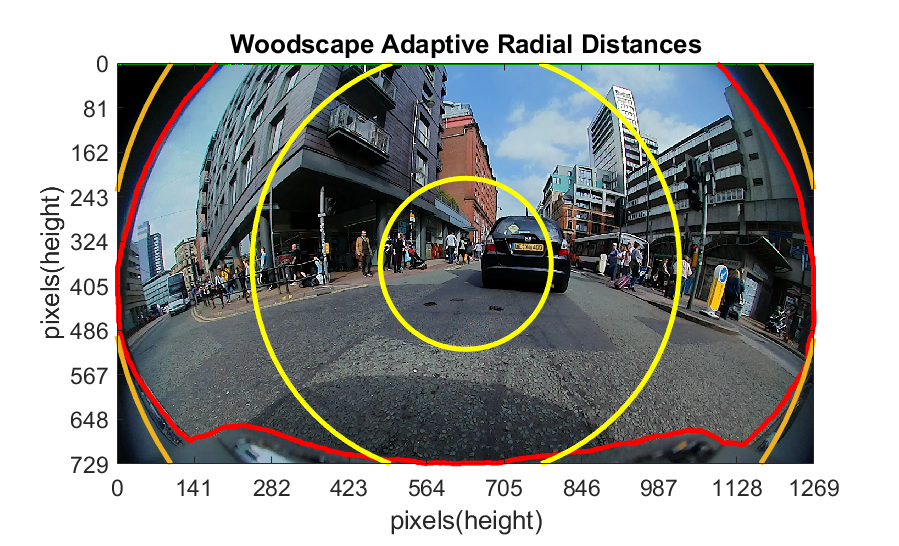}}
    \hspace{1cm}
\subfloat[SynWoodscape\label{fig:synwood}]{\includegraphics[width=2.5in, keepaspectratio]{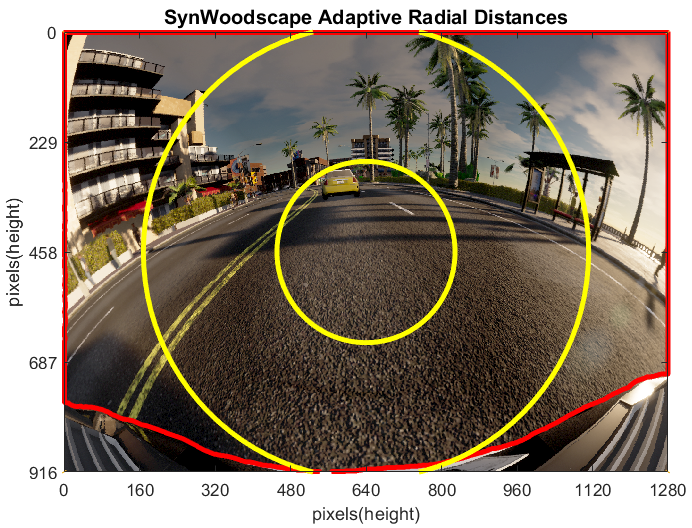}}
    \caption{\emph{\textbf{Adaptive Radial Distances for all four datasets:} (a) KITTI (no mask), (b) LMS (circular mask), (c) Woodscape (Front View mask) and (d) SynWoodscape (Front View mask). Note: all masks are marked by a red outline, inner radial distances (yellow circles) and the outer radial distance (orange circle). For KITTI only the inner Radial Distances are shown between (200-1200) pixels.}}
    \label{fig:nssfr-custom-masks}
\end{figure*}

\label{sec:cust-roi}
To generalize processing lenses for NS-SFR, the main interest is to use the natural scene captured by each camera lens. To do this, certain optical effects especially in wide FOV cameras need to be removed which are not part of the natural scene. This involves removing effects such as mechanical vignetting (part of the camera itself is visible in the corners of fisheye images) and ego-vehicle occlusion. These can bias the measurements of NS-SFR where the part of the car to which the camera is attached is visible in any image. All custom masks are created using the Freehand ROI tool from MATLAB\footnote[1]{See Mathworks for information on Freehand ROI: \url{https://uk.mathworks.com/help/images/ref/drawfreehand.html}} allowing the user to customize a particular ROI mask to isolate the natural scene. If a new camera is used, a new mask is created to fit the scene(see Fig. \ref{fig:nssfr-custom-masks}). %The KITTI dataset was the exception, as there was no need for a mask due to the absence of ego-body occlusion and mechanical vignetting.
\subsection{Adaptive Radial Distances}
To further analyze the images three adaptive radial distances were created which categorize the location of each slanted edge according to its relative distance from the center of the image. The technique used to create each segment is by taking the Euclidean distance from the central point to the farthest edge of the custom ROI mask. In the case of KITTI, the corner of the image was taken as the farthest distance from the center of the image. The Euclidean distance calculation is represented by the following formula: 
%\begin{equation}
%    \begin{aligned}
%    r_e = \left\{ \begin{array}{cl}
%    max\left(\sqrt{(x_c - x_m)^2 + (y_c - y_m)^2}\right), & (x_m,y_m)\in[X_m, Y_m] \\
%    \sqrt{(x_c - x)^2 + (y_c - y)^2}, & else
%    \end{array} \right. \nonumber
%    \end{aligned}
%    \label{eq:blue-rad-dist}
%\end{equation}
\vspace{-0.3cm}
\begin{equation}
    r_e =  \max\left(\sqrt{(x_c - x_m)^2 + (y_c - y_m)^2}\right) \forall (x_m,y_m)\in[X_m, Y_m]
    \label{eq:blue-rad-dist}
\end{equation}
Where, $r_e$ is the Euclidean distance representing the radius of the largest radial distance (i.e. the orange circles (with the exception of KITTI where the orange circle is not visible due to the radius being the distance from corner to centre of the image) in Fig. \ref{fig:nssfr-custom-masks}), the point ($x_c$, $y_c$) represent the location of the centre of the image or mask, ($x_m$, $y_m$) represent points along the periphery of the custom ROI mask $[X_m, Y_m]$ and can be the location of the bottom right corner of the image if no mask is present. (Note: In the case that no ROI is set, $[X_m, Y_m]$ is the periphery of the image). The radii of the radial distances/annuli are obtained by dividing the maximum radius $r_e$ into $N$ proportional parts.
%The radiuses of each of the annuli are calculated by multiplying the radius of the largest radial distance by a ratio. Let $r_3 = r_e$ be the radius of the outer radial distance where $i > 1$ and $i = i - 1$:
%\begin{equation}
%    \begin{aligned}
%        r_i = r_i * \frac{n}{(N + 1)}
%    \end{aligned}
%    \label{eq:rad-dist-ratio}
%\end{equation}

%Where N is the number of four radial segments (the fourth one is ignored) and n is the index between the middle two radii plus one. Hence the radius of the blue circle is multiplied by 0.6 and 0.4 ratios to obtain the \nth{2} and \nth{1} radial distances (black circles) respectively. Another way of observing this is that the outer radius of the \nth{3} radial distance (blue circle) is approximately four times that of the central radial distance.
The ratios determined in these calculations are suitable mainly because they give sufficient segments to divide up the images into three categories (i.e., center, middle, and edge) and generalize well regardless of camera type. Ratios smaller than this are unsuitable (i.e., there tends to be bias towards having too many slanted edges favored for one segment).
\begin{figure*}
    \centering
    \captionsetup{font={normalsize}, textfont=sf} 
    \subfloat[Valid and Invalid ROI Selection \label{fig:wood-roi}]{\includegraphics[width=3.5in, keepaspectratio]{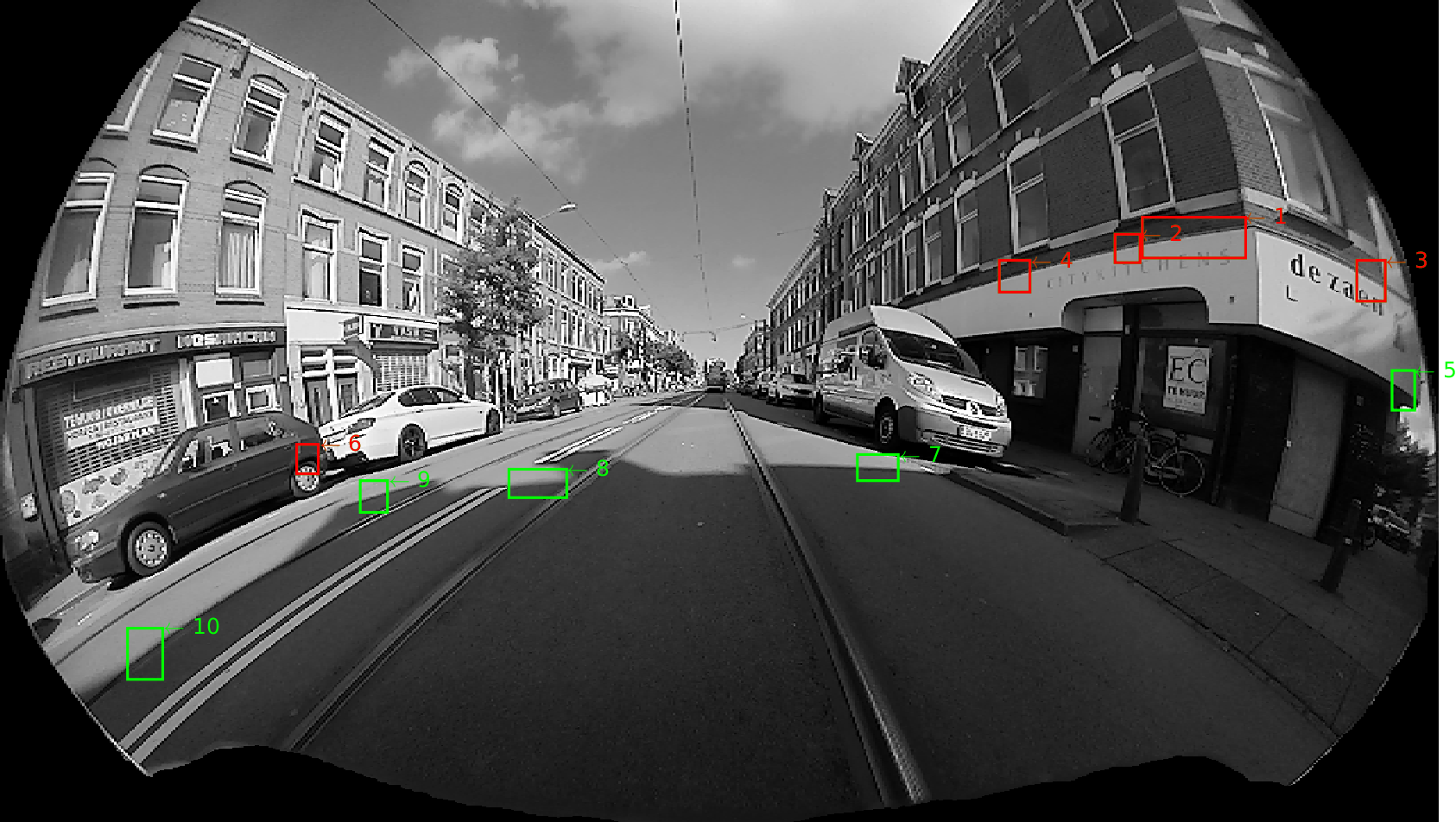}}  \hspace{1cm}
    \subfloat[Measurements \label{fig:wood-meas}]{\includegraphics[width=2.1in, keepaspectratio]{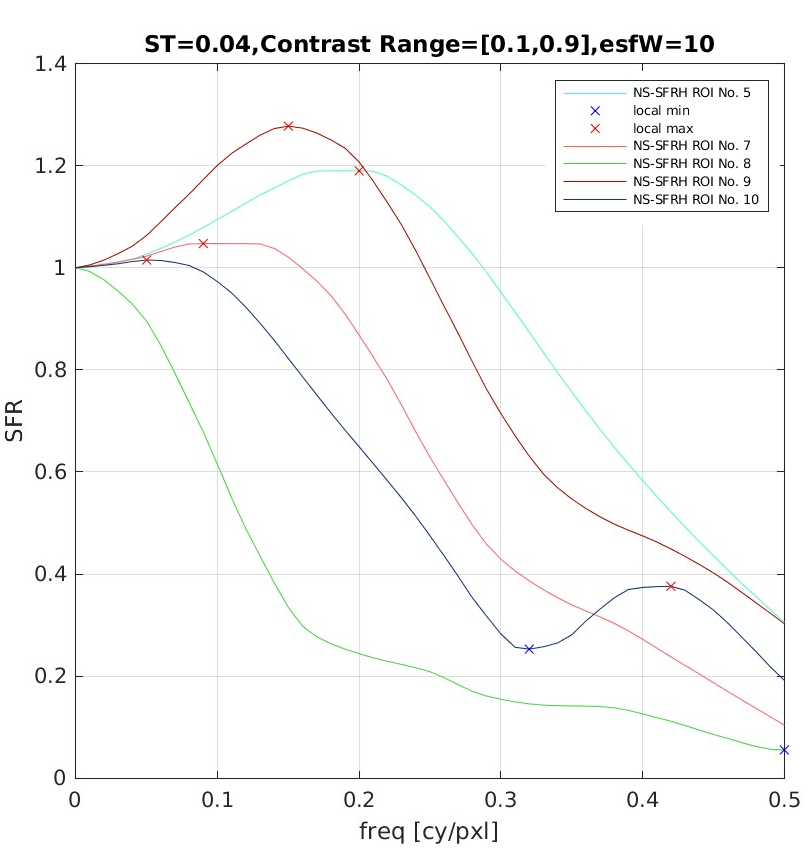}}
    \caption{\emph{An example of \textbf{ROI Selection} from one of the Front View of the Woodscape dataset at (ST=0.04, esfW=10). Observe that the local maxima (red) and local minima (blue) of each measurement are marked by an x symbol.}}
    \label{fig:Wood-roi-select}
\end{figure*}

%\begin{figure}[t]
%    \centering
%    \includegraphics[width=2in, keepaspectratio]{results/Wood/legend.jpg}
%    \caption{Legend for Fig. \ref{fig:Wood-roi-select}}
%    \label{fig:KITTI-roi-res}
%\end{figure}

\subsection{Re-use Optimized NS-SFR}
The ROI selection technique and parameter tuning of NS-SFR optimized algorithm \cite{van2023tool} is reused here. Canny Edge Detection with edge masking is used to find edges from the natural scenes and isolate edges of interest \cite{van2023tool}. Unlike measurement test charts, NS-SFR does not have a strong sharpening effect on step edges due to the surrounding scene content \cite{van2023tool}.
%Pixel-stretching for finding suitable edges from the datasets as well as frequency, noise, pixel contrast, and slanted edge angle are all used. %This strategy is re-used in the following experiments to select slanted edges from images.
The NS-SFR parameters that determine slanted edge selection in natural scenes are the following (the parameters used are given):
\begin{enumerate}
    \item{\textbf{Contrast Range} (0.1 - 0.9) - the contrast range between the transition from dark to white in a region with a slanted edge:
    \begin{enumerate}
    \item{Low contrast ($<$0.1)  is prone to noise error,}
    \item{High contrast ($>$0.9) is prone to non-linear sharpening and image processing,}
    \end{enumerate}
    }
    \item{\textbf{Edge Angle Range} (\ang{0} - \ang{360}) - there is no restriction on the angle range of slanted edges as it limits the number of edges that can be found in a natural scene except for \ang{0} and \ang{45} which are filtered out and is not possible due to errors \cite{ISOBS2017},}
    \item{\textbf{Step Edge Noise Floor (ST)} (0.02 and 0.04 (for images with greater noise))  ensures that the gradient on either side of the edge is uniform. For example, a value of 
    0.02 signifies a change of pixel value or digital number (DN) of 4.5 for an 8-bit image and 0.04 is twice this value at 9 \cite{van2021natural}. It was observed that sharper edges were found as the ST and esfW parameters increased,}
    \item{\textbf{Edge Spread Function Width (esfW)} (default value is $5$ pixels) remove edges too close together. For example, edges can be 5 pixels apart in an image,}
\end{enumerate}
For initial experiments, a few default parameters were used (i.e., ST=0.02, esfW=5 pixels) \cite{van2021natural}. These were found to be especially suitable for narrow FOV cameras, such as KITTI. The only difference to the default was removing the Edge Angle range and increasing the default contrast range between (0.55-0.65) to (0.1-0.9) to accommodate for a larger ROI selection. As a second experiment, the default parameters were increased to (ST=0.04,esfW=10 pixels) which showed a higher quality ROI selection for wide FOV cameras (particularly for Woodscape).
%However, a trial-and-error basis is employed for varying the parameters supplied to the NS-SFR algorithm. The parameter combination that gives a reasonable number of good-quality images is identified for each dataset.
This was done to understand what happens if distortion is applied to the images and whether the algorithm performs better with a different parameter combination than the default used in the experiments in the prior art \cite{van2022camera}.
In the end, we settle on a unique parameter combination for each dataset where each radial segment of the spatial domain is well represented (i.e., each segment should have more than 20 valid slanted edges to calculate the mean).
\subsection{Valid \& Invalid ROI Selection}
We apply additional constraints to slanted edge selection where the following constraints were applied:

\begin{enumerate}
    \vspace{-0.3cm}
    \item{\textbf{Local Maxima $<$ 1.4 SFR} - If the local maxima exceed a peak of 1.4 SFR ($\cong25\%$ overshoot), the measurement is discarded. 25\% overshoot or less is acceptable\footnote[2]{See Imattest website for information on MTF curves and Image appearance: \url{https://www.imatest.com/support/docs/23-1/mtf_appearance/}} for slanted edges,}
    \item{\textbf{Local Minima $<$ 0.4 SFR} - If the local minima exceed 0.4 SFR, the measurement is discarded. Setting this would filter out signals with excessive noise past the Nyquist frequency (0.5cy/px) as shown in Oliver et al. \cite{van2019edge},}
    \vspace{-0.3cm}
\end{enumerate}
Measurements that meet the above requirements are marked by a green bounding box and those that do not are marked by a red bounding box as illustrated in Fig. \ref{fig:wood-roi}.
\subsection{Radial Distance Analysis}
%A new approach is proposed to analyze slanted edges.
Instead of using three evenly distributed radial distances \cite{van2023tool}, three adaptive radial distances are proposed dependent on the regional mask. The mean NS-SFR per radial distance is calculated and compared visually between all four datasets. The mean NS-SFR are plotted against each other to gain insight into how the varying spatial domain affects image quality.
\begin{figure*}[t]
    \centering
    \captionsetup{font={normalsize}, textfont=sf}
    \hspace{-1cm}
    \subfloat[KITTI (ST = 0.02, esfW = 5)\label{fig:KITTI-spa-dist}]{\includegraphics[width=2.5in, keepaspectratio]{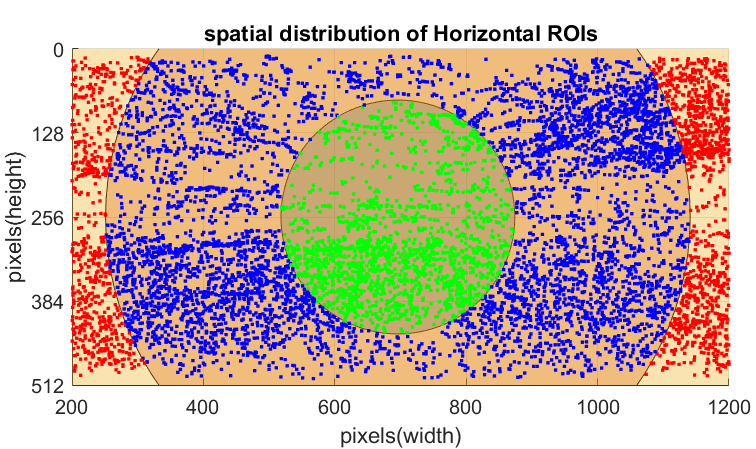}}
    \hspace{1cm}
    \subfloat[LMS (ST = 0.02, esfW = 5)\label{fig:lms-spa-dist}]{\includegraphics[width=2.2in, keepaspectratio]{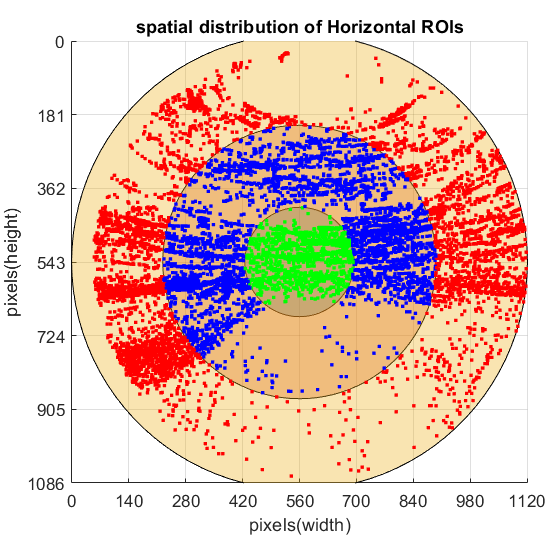}}\\
    \subfloat[Woodscape (ST = 0.04, esfW = 10)\label{fig:wood-spa-dist}]
    {\includegraphics[width=2.5in, keepaspectratio]{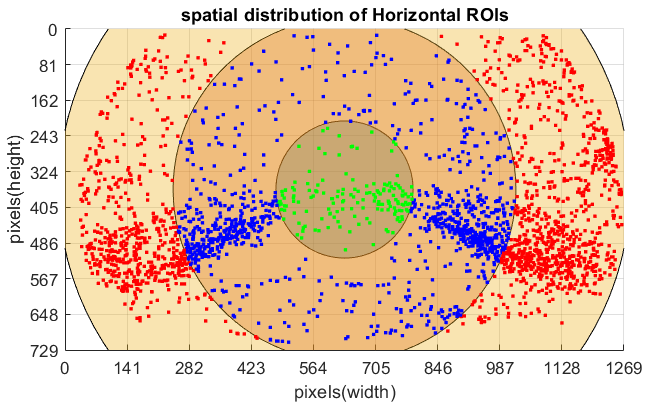}}
    \hspace{1cm}
    \subfloat[SynWoodscape (ST = 0.04, esfW = 10)\label{fig:synwood-spa-dist}]{\includegraphics[width=2.6in, keepaspectratio]{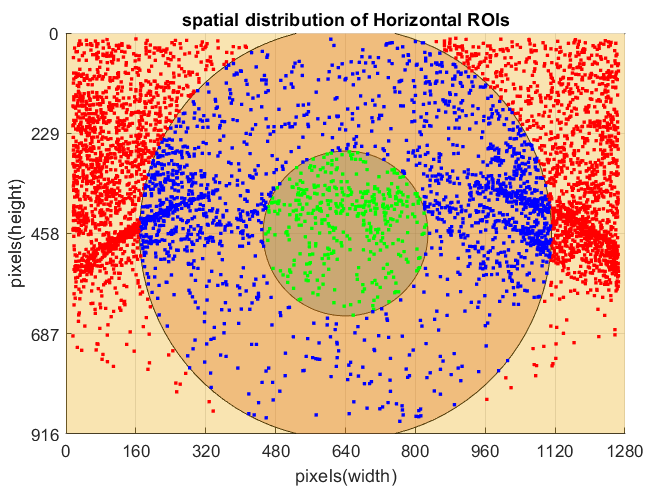}}
    \caption{\emph{\textbf{Horizontal Slanted Edge Locations of all four datasets (refer to Fig. \ref{fig:nssfr-custom-masks} for Adaptive Distances):}  (a) KITTI, (b) LMS, (c) Woodscape and (d) SynWoodscape. Distributions show the pattern of ROI selection for NS-SFR for both narrow and wide FOV cameras. Here, we can see the trend of the location of slanted edges in the scenes for the given datasets.}}
    \label{fig:spa-dist}
    \vspace{-0.3cm}
\end{figure*}

\section{Results}
\subsection{Datasets}
The datasets used in this study are:
\begin{enumerate}
\vspace{-0.2cm}
\item{\textbf{1065 Front View 90° FOV KITTI city images \cite{geiger2013vision}},}
\item{\textbf{1251 LMS 185° FOV real-life images from Drive A \cite{eichenseer2016data}},}
\item{\textbf{1514 Front View 190° FOV Woodscape images \cite{Yogamani2019}},}
\item{\textbf{500 Front View images of SynWoodscape \cite{sekkat2022synwoodscape}},}
\vspace{-0.2cm}
\end{enumerate}
Each dataset has one lens calibration and a custom ROI mask applied to the images.
The first experiment is to demonstrate the operation of the NS-SFR algorithm on a dataset with little to no distortion (i.e., the KITTI dataset \cite{geiger2013vision}). No mask is applied to these images.
The second experiment is to use a circular fisheye dataset from \cite{eichenseer2016data} with wide FOV. A custom circular mask is applied to these images.
For the third experiment, a typical automotive dataset, the Woodscape dataset \cite{Eising2021} is used in this work for wide FOV (i.e., 190° field-of-view) cameras.
Finally, SynWoodscape is used which is a synthetic version of Woodscape created in the CARLA simulator using 4th-order polynomial cube-map projection \cite{sekkat2022synwoodscape, dosovitskiy2017carla}.

\subsection{KITTI camera experiments}
%To demonstrate that the proposed pipeline is functional, adapted NS-SFR was applied on \ang{90} FOV KITTI images. A sample scene is shown in Fig. \ref{fig:KITTI-roi-select} where the green boxes represent valid slanted edge measurements with minimal edge enhancement and noise. As can be seen, NS-SFR has identified 82 potential edges of which only 21 were chosen as valid which is approximately 25.6\%. Considering the amount of noise natural scenes tend to have it can be concluded that KITTI has a dense selection of slanted edges. The results of the ROI selection from Fig. \ref{fig:KITTI-roi-select} is shown in Fig. \ref{fig:KITTI-roi-res} where all 21 slanted edges are graphed.
ROI selection was iterated over 1065 images of a natural scene in the KITTI dataset and a resulting spatial distribution of the slanted edges can be seen in Fig. \ref{fig:KITTI-spa-dist}. There was no adaptive regional mask required for KITTI. The yellow radial distances from Fig. \ref{fig:KITTI} were calculated by taking one of the corners of the KITTI images which is by default the farthest location away from the centre of each image and the radiuses were determined by a ratio of $N$ proportional segments. The mean MTF of each segment is shown in Fig. \ref{fig:KITTI-mean-mtf}.
\subsection{The LMS Circular fisheye camera experiments}
The experiments were repeated on 185° FOV circular fisheye images. A circular mask was created for the dataset illustrated in Fig. \ref{fig:lms} where the blue radial distance is calculated based on the furthest point of the circular mask using (\ref{eq:blue-rad-dist}). Fig. \ref{fig:lms-spa-dist} represents the resulting spatial distribution of the LMS dataset subsample used in the experiments. In comparison to KITTI, this has a circular aperture and a completely different ROI distribution. The default parameters were used in the KITTI dataset.
The mean MTF of each segment is shown in Fig. \ref{fig:lms-mean-mtf} where all MTF50 measurements are worse than in KITTI (see Fig. \ref{fig:KITTI-mean-mtf}).
This is the first dataset to exceed the \ang{180} FOV in these experiments.
\subsection{Woodscape experiments}
The Woodscape dataset is composed of \ang{190} FOV cameras \cite{Eising2021}. Out of the four camera perspectives available, the Front View perspective was chosen for experiments that consisted of 1514 images.
To demonstrate the proposed pipeline on wide FOV, a sample slanted edge ROI selection is shown in Fig. \ref{fig:Wood-roi-select} where the green boxes represent valid slanted edge measurements with minimal edge enhancement and noise as specified by the local maxima and minima constraints. As can be seen, NS-SFR has identified 10 potential edges of which only 5 were chosen as valid which is approximately 50\%. Considering the amount of noise natural scenes tend to have it can be concluded that Woodscape has good selection of slanted edges. Slanted Edge measurements of each ROI from Fig. \ref{fig:wood-roi} are shown in Fig. \ref{fig:wood-meas} where all 5 slanted edges are graphed.
The ROI mask of Woodscape is shown in Fig. \ref{fig:wood} where the images had both ego-vehicle body occlusion and mechanical vignetting. The experiments resulted in another unique spatial distribution shown in Fig. \ref{fig:wood-spa-dist}. For this experiment, both ST and esfW were increased to 0.04 and 10 respectively. The mean MTF of each segment is shown in Fig. \ref{fig:wood-mean-mtf}. %This is the second fisheye lens investigated in these experiments.
\subsection{SynWoodscape experiments}
\begin{figure*}[t]
    \centering
    \captionsetup{font={normalsize}, textfont=sf} 
    \subfloat[KITTI  (ST = 0.02, esfW = 5)\label{fig:KITTI-mean-mtf}]{\includegraphics[width=2.6in, keepaspectratio]{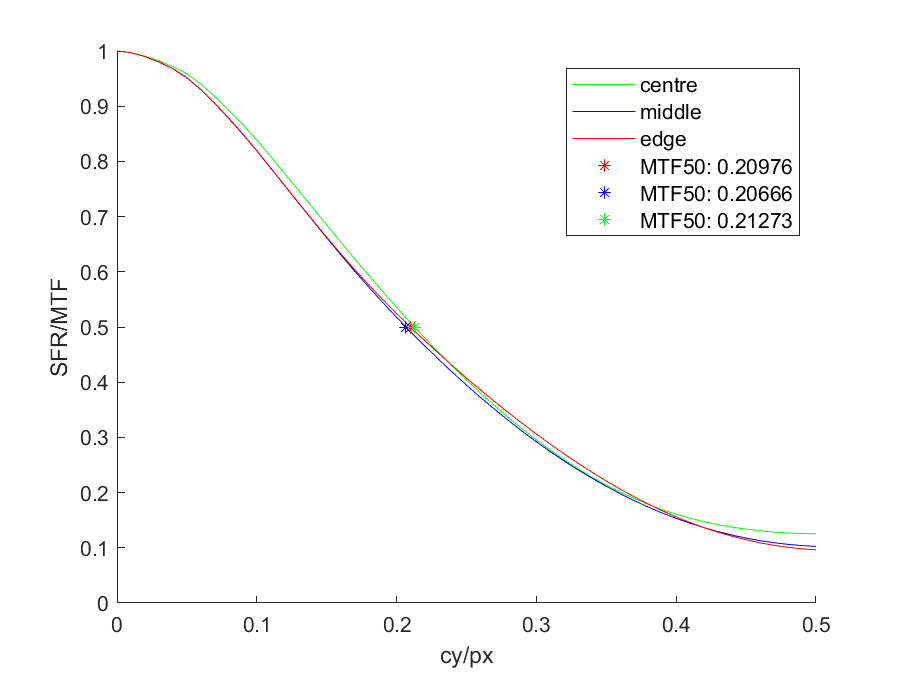}}
    \qquad
    \subfloat[LMS  (ST = 0.02, esfW = 5)\label{fig:lms-mean-mtf}]{\includegraphics[width=2.6in, keepaspectratio]{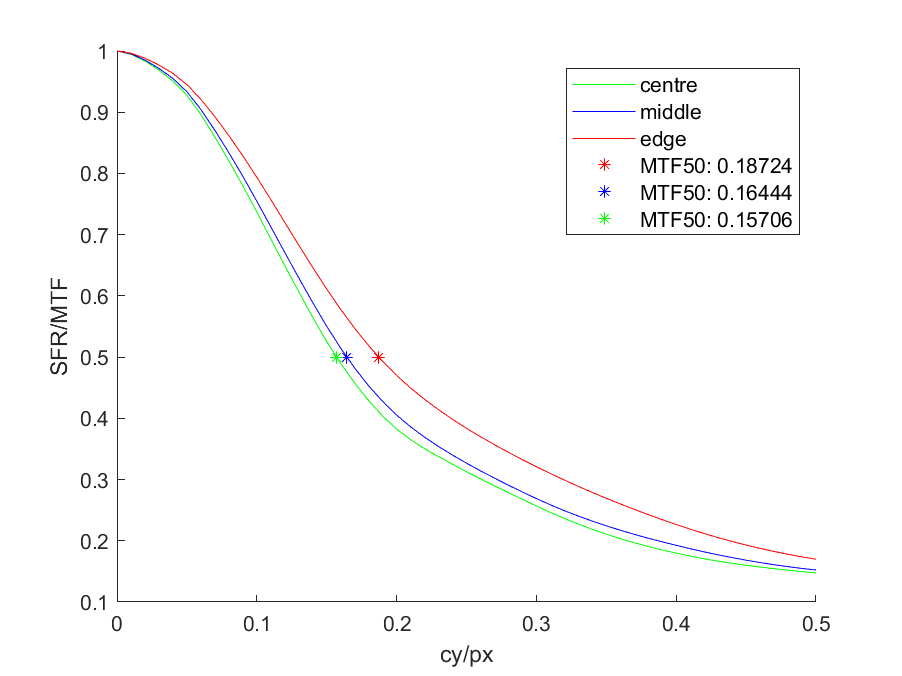}}\\
    \subfloat[Woodscape  (ST = 0.04, esfW = 10)\label{fig:wood-mean-mtf}]{\includegraphics[width=2.6in, keepaspectratio]{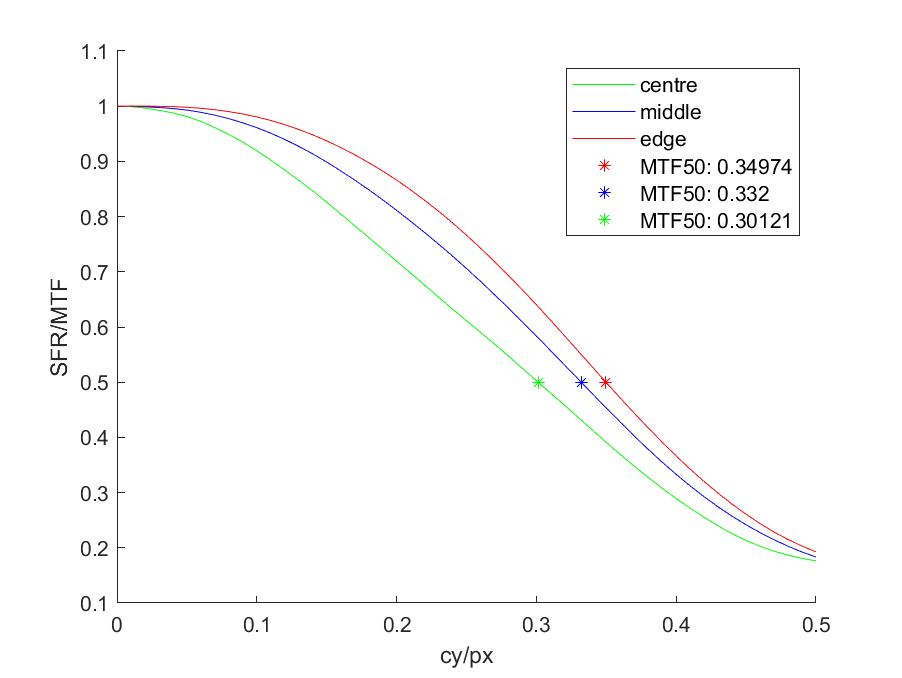}}
    \qquad
    \subfloat[SynWoodscape  (ST = 0.04, esfW = 10)\label{fig:synwood-mean-mtf}]{\includegraphics[width=2.6in, keepaspectratio]{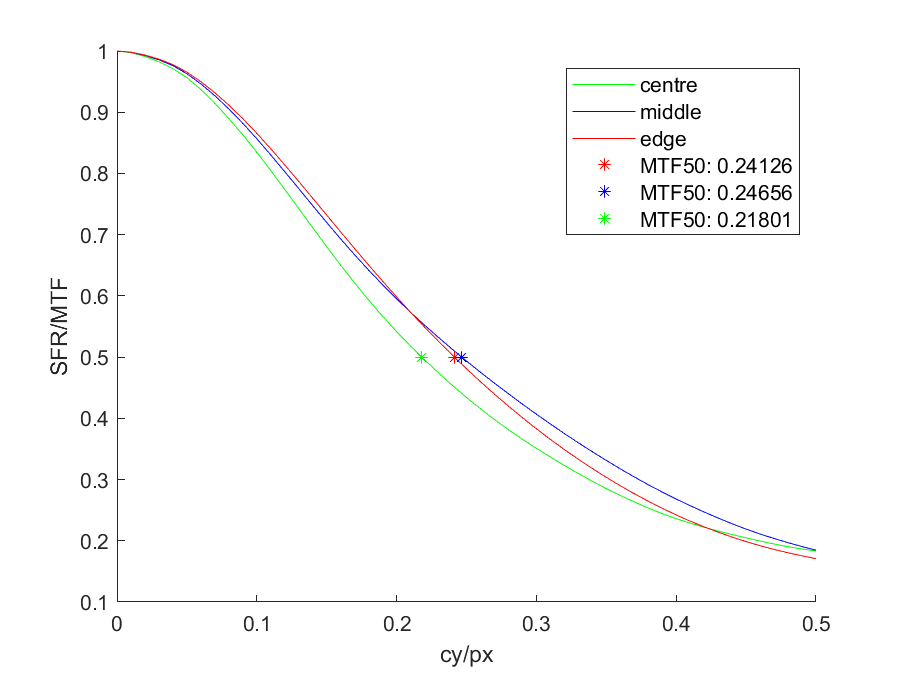}}
    \caption{\emph{NS-SFR MTF50 Results of four automotive datasets}.}
    \label{fig:mtf50-res}
\end{figure*}

For SynWoodscape experiments, the custom ROI mask can be seen in Fig. \ref{fig:synwood}. Unlike Woodscape SynWoodscape does not have dark corners from mechanical vignetting and only ego-body occlusion needs to be removed from the images. The spatial distribution of this experiment can be seen in Fig. \ref{fig:synwood-spa-dist}. By comparing Fig. \ref{fig:synwood-spa-dist} to Fig. \ref{fig:wood-spa-dist}, SynWoodscape has a much denser and wider spread of points than Woodscape using the same parameter settings despite having only 500 images to work with. This comparison proves that in a perfect world simulated scenes tend to have more ideal slanted edges than in real life. For consistency, the same NS-SFR parameters as for Woodscape were used. The mean MTF of each segment is shown in Fig. \ref{fig:synwood-mean-mtf}.
\subsection{Discussion}
In Fig. \ref{fig:spa-dist} we show the spatial distribution of slanted edges found by NS-SFR and we can observe distinct patterns emerging where scene dependence affects ROI selection in images with greater distortion. For example, the selection of slanted edges for KITTI Fig. \ref{fig:KITTI-spa-dist} is mostly uniform whereas for wide FOV, ROI selection follows the distortion of the natural scenes. The LMS dataset has a significant number of slanted edges identified on the buildings or in the sky (see Fig. \ref{fig:lms-spa-dist}) whereas, for both Woodscape and SynWoodscape, clusters of slanted edges are found along the horizon where the road meets the sky and the buildings (see Fig. \ref{fig:wood-spa-dist} and \ref{fig:synwood-spa-dist}). However, a greater density of slanted edges does not imply better quality selection as is evident in the LMS MTF50 results in Fig. \ref{fig:lms-mean-mtf} where the quality of slanted edge selection is poor (mean MTF50 is between 0.15-0.18 cy/px). Parameter variations visibly provide higher quality but fewer slanted edge selections for Woodscape (as shown in Fig. \ref{fig:wood-spa-dist}) whereas the periphery provides better-slanted edge selection than the centre as shown in Fig. \ref{fig:wood-mean-mtf}. In contrast to LMS, Woodscape has better MTF50 between (0.31-0.35 cy/px). The SynWoodscape dataset has lower-quality slanted edges than Woodscape suggesting that simulation does not imply higher-quality slanted edges.
The central radial segment has the least sharp slanted edges, the middle segment is the second sharpest and finally, the third radial segment has the sharpest edges. Peter Burns et al \cite{burns2020application}, shows \nth{3} order and \nth{5} order polynomial measurements on curved edges with similar results. All measurements are of \nth{5} order in these natural scenes.
\section{Future Work}
\begin{itemize}
    \item Consider automated regional masking for each fisheye perspective as originally proposed by Hogan et al. \cite{hogan2022automatic}.
    \item Differentiate SFR of different objects in natural scenes.
\end{itemize}
\section{Conclusion}
In this paper, we have demonstrated the adaptation of NS-SFR to automotive datasets. We have demonstrated a novel approach to isolate natural scenes by applying regional masking (where required) and measuring image quality using an Adaptive Radial Distance Analysis on four datasets (i.e., KITTI, LMS, Woodscape and SynWoodscape). It was found that Woodscape provided the best selection of slanted edges where MTF50 measurements are the highest out of the four datasets. For horizontal edges, image quality degrades at the centre of the image. 
\section{Acknowledgments} 
%add the acknowledgement section here
This work was supported, in part, by the Science Foundation
Ireland grant 13/RC/2094\_P2 and co-funded under the European Regional Development Fund
through the Southern \& Eastern Regional Operational Programme to Lero - the Science
Foundation Ireland Research Centre for Software (www.lero.ie). The authors of this work would like to thank Oliver van Zwanenberg for assisting in the adaptation of the NS-SFR framework.
%%%%%%%%%%%%%%%%%%%%%%%%%%%%%%%%%%
% Bibliography
%%%%%%%%%%%%%%%%%%%%%%%%%%%%%%%%%%
\bibliographystyle{IEEEtran}
\small
\bibliography{ieee-references}
\end{document}